\documentclass[runningheads]{llncs}
\usepackage[T1]{fontenc}

\usepackage{graphicx}
\usepackage{color}
\usepackage{hyperref}
\usepackage{amsmath,amssymb,amsfonts}
\usepackage{algorithmic}
\usepackage{graphicx}
\usepackage{textcomp}
\usepackage{booktabs} 
\usepackage{placeins}
\usepackage{xcolor}
\usepackage{multirow}
\usepackage{wrapfig}
\definecolor{PineGreen}{RGB}{1, 121, 111}

\def\BibTeX{{\rm B\kern-.05em{\sc i\kern-.025em b}\kern-.08em
    T\kern-.1667em\lower.7ex\hbox{E}\kern-.125emX}}

\newcommand{\shenc}{\mathcal{E}_\text{shared}}
\newcommand{\bh}{\mathbf{h}}

\newcommand{\be}{\mathbf{e}}
\newcommand{\bv}{\mathbf{v}}
\newcommand{\bz}{\mathbf{z}}

\begin{document}
\title{A Shared Encoder Approach to Multimodal Representation Learning}

\author{Shuvendu Roy\textsuperscript{1,2}~ Franklin Ogidi\textsuperscript{1}~ \\Ali Etemad\textsuperscript{2}~ Elham Dolatabadi\textsuperscript{1,3}~ Arash Afkanpour\textsuperscript{1}\thanks{Corresponding author. Email: arash.afkanpour@vectorinstitute.ai}\\
$^1$Vector Institute~~~
$^2$Queen's University, Canada~~~
$^3$York University, Canada
}
\institute{}
\maketitle              

\begin{abstract}
Multimodal representation learning has demonstrated remarkable potential in enabling models to process and integrate diverse data modalities, such as text and images, for improved understanding and performance. While the medical domain can benefit significantly from this paradigm, the scarcity of paired multimodal data and reliance on proprietary or pretrained encoders pose significant challenges. In this work, we present a shared encoder framework for multimodal representation learning tailored to the medical domain. Our approach employs a single set of encoder parameters shared across modalities, augmented with learnable modality features. Empirical results demonstrate that our shared encoder idea achieves superior performance compared to separate modality-specific encoders, demonstrating improved generalization in data-constrained settings. Notably, the performance gains are more pronounced with fewer training examples, underscoring the efficiency of our shared encoder framework for real-world medical applications with limited data. Our code and experiment setup are available at \href{https://github.com/VectorInstitute/shared_encoder}{https://github.com/VectorInstitute/shared\_encoder}.

\end{abstract}

\section{Introduction}
Multimodal representation learning \cite{clip,ALIGN} has emerged as a powerful paradigm for integrating diverse data sources, such as text, images, audio, and video, to enable more comprehensive and robust models. This approach has driven progress in fields like natural language processing, computer vision, and vision-language modeling. In the medical domain, multimodal learning holds particular promise, as medical data often comes in various forms, including radiology images, clinical notes, and Electronic Health Records (EHR). Effectively combining these modalities can improve diagnostic accuracy, enhance clinical decision support, and enable more holistic patient assessments.

However, learning robust multimodal representations in the medical domain presents unique challenges. A key limitation is the scarcity of paired data across modalities, which many multimodal learning approaches rely on for contrastive or fusion-based training \cite{clip,ALIGN,girdhar2023imagebind}. Unlike general-purpose domains where internet-scale annotated corpora enable training large-scale multimodal models, medical data is often fragmented, privacy-restricted, and labor-intensive to annotate. This data scarcity makes it difficult to train modality-specific encoders that generalize well across medical tasks, particularly when large-scale pretraining is not feasible.

To address these challenges, we propose a shared encoder framework for multimodal representation learning. Instead of training separate encoders for each modality, our approach leverages a single encoder shared across modalities, enabling more efficient and unified representation learning. This design has several advantages, particularly in low-data settings: (1) parameter sharing across modalities reduces the risk of overfitting and improves generalization, (2) the encoder can leverage shared structures and relationships present in medical data, and (3) it allows for more efficient adaptation to new modalities or tasks with limited labeled examples.

To enhance the flexibility of the shared encoder, we incorporate modality-specific information through learnable modality embeddings, which help the model distinguish between different input modalities while still benefiting from shared feature extraction. We systematically investigate the effectiveness of this strategy and explore further architectural refinements, such as incorporating modality-specific layers before or after the shared encoder to balance specialization and generalization.

We evaluate our approach through extensive experiments. Specifically, we pretrain our shared encoder on the PMC-OA dataset \cite{lin2023pmc} and assess its performance on both in-distribution and out-of-distribution retrieval tasks. Our results show that the shared encoder outperforms traditional modality-specific encoders, particularly in low-data settings, and that performance can be further improved by introducing modality-specific embeddings and early specialization layers. Our main contributions are as follows:
\begin{enumerate}
\item We introduce a shared encoder framework for multimodal representation learning, demonstrating its effectiveness in low-data scenarios, particularly in the medical domain.
\item We conduct a systematic study on the impact of modality-specific information in a shared encoder setup, showing significant performance gains when using learnable modality embeddings.
\item We explore architectural refinements, such as early modality-specific layers, and demonstrate their benefits in balancing generalization and specialization.
\item We will release our implementation to facilitate reproducibility and further research in multimodal medical learning.
\end{enumerate}

\section{Related Work}
Contrastive vision-language learning has become a prominent area of research in self-supervised learning, where models align visual and textual modalities using contrastive loss to learn generalizable features from large datasets without supervision. CLIP \cite{radford2021learning} is a seminal work demonstrating that task-agnostic pretraining can achieve performance comparable to task-specific supervised models. ALIGN \cite{jia2021scaling} further extended this approach by scaling both data and model size, utilizing over a billion noisy image-text pairs. However, the substantial computational and data requirements of these models have led to efforts focused on more efficient pretraining methods. DeCLIP \cite{li2021supervision} enhances training on smaller datasets with nearest-neighbour supervision, while LiT \cite{zhai2022lit} reduces computational costs by employing a frozen image encoder. 

In the medical domain, vision-language models show promise for applications like chest X-rays \cite{boecking2022making,tiu2022expert,seibold2022breaking}, ECG analysis \cite{liu2024etp}, histopathology \cite{zhang2023text,ikezogwo2024quilt}, lung CT scans \cite{liu2023imitate}, dermatological imaging \cite{kim2023concept}, and eye fundus photography \cite{baliah2023exploring}. These models face challenges related to the scarcity of annotated data and the distinctive nature of medical images, where even small abnormalities are diagnostically significant. To mitigate data scarcity, several studies have explored generating synthetic image-text pairs across medical domains \cite{lin2023pmc,eslami2021does,zhang2023biomedclip}. Fine-grained approaches like GLoRIA \cite{huang2021gloria} and LoVT \cite{muller2022joint} improve cross-modal learning by focusing on local-level interactions between textual descriptions and specific image regions, crucial for diagnostic understanding. Similarly, CONCH \cite{lu2024visual} integrates generative loss to capture region-specific features, benefiting tasks like visual recognition and image captioning.

Few studies have explored sharing parameters between encoders. Most rely on a pretrained encoder as the shared component, with some introducing modality-specific parameters. For example, VIT-LENS \cite{lei2024vit} uses a pretrained ViT encoder in a contrastive learning framework for additional modalities. Other approaches partially share encoder parameters, like ONE-PEACE \cite{wang2305one}, where only 378 million out of 4 billion parameters are shared across modalities. FLAVA \cite{singh2022flava} combines modality-specific encoders with a shared, modality-agnostic encoder. Gato \cite{reedgeneralist} presents a generalist agent that uses a single transformer to handle diverse tasks and modalities, including text, images, and robotic control. HighMMT \cite{lianghigh} proposes a multimodal model that generalizes across high-modality and partially observable scenarios, using a shared encoder for tasks involving text, images, audio, and other data types. MoMo \cite{chada2023momo} introduces a shared encoder model for text, image, and multimodal representations, achieving strong performance while being data, memory, and runtime efficient. In contrast to these works, our work aims to develop a fully shared multimodal encoder that enhances representation learning for all modalities.

\section{Methodology}
\label{sec:methodology}
\subsection{Preliminaries}

Vision-language pretraining has become a cornerstone of multimodal representation learning, enabling the development of models that align visual and textual modalities. One of the most prominent frameworks for this purpose is CLIP (Contrastive Language-Image Pretraining) \cite{clip}, which employs contrastive learning to maximize the similarity between paired image-text representations and minimize the similarity of mismatched pairs. Given an image $I$ and its corresponding caption $T$, CLIP uses separate encoders for images and text to generate latent representations $\bz_I$ and $\bz_T$. The training objective is defined as:
\begin{equation}
\small
    \mathcal{L}_{\text{con}} = - \frac{1}{N} \sum_{i=1}^{N}  \left( \log \frac{\exp (\langle\bz_I^i, \bz_T^i \rangle /\tau)}{\sum_{j=1}^{N} \exp (\langle\bz_I^i, \bz_T^j \rangle / \tau)}   +  \log \frac{\exp(\langle \bz_T^i, \bz_I^i \rangle/\tau)}{\sum_{j=1}^{N} \exp (\langle\bz_T^i, \bz_I^j \rangle /\tau)} \right),
\label{eq:contrastive}
\end{equation}
where $\langle\cdot, \cdot \rangle$ denotes the cosine similarity, $\tau$ is a temperature parameter, and $N$ is the batch size.

To form the input to a transformer-based encoder we follow the standard multimodal representation learning techniques like CLIP, i.e., each example is first converted into a sequence of length $s$, e.g., by tokenizing text or splitting an image into patches. Each element in this sequence is then mapped to a $d$-dimensional vector representation using a modality-specific embedding module.
In our framework since all input vectors are processed by a single encoder, the dimensionality of input representation vectors ($d$) must be consistent across modalities. The resulting sequence of embedding vectors serves as the input to the encoder. Let $[\be_M^1, \ldots, \be_M^s]$ denote the sequence of length $s$ representing the input vectors for modality $M$.

Contrastive learning has proven effective for cross-modal alignment in general-domain tasks, where large paired datasets are available. However, its application in the medical domain is limited by data scarcity. One approach to tackle this problem is controlling model complexity by reducing the number of learnable parameters, which we will discuss in the next section.

\subsection{Shared Encoder Framework}
To address the challenges of limited data and modality-specific dependencies, we propose a shared encoder framework in which most parameters are shared across modalities. Figure~\ref{fig:main} (top) illustrates the concept of a shared encoder at a high level.

Our framework consists of two components: a \emph{modality identifier} and a \emph{shared encoder}.
We describe each component in detail below.

\subsubsection{Modality Identifier.}
\begin{figure*}[tb]
    \centering
    \includegraphics[width=1.0\linewidth]{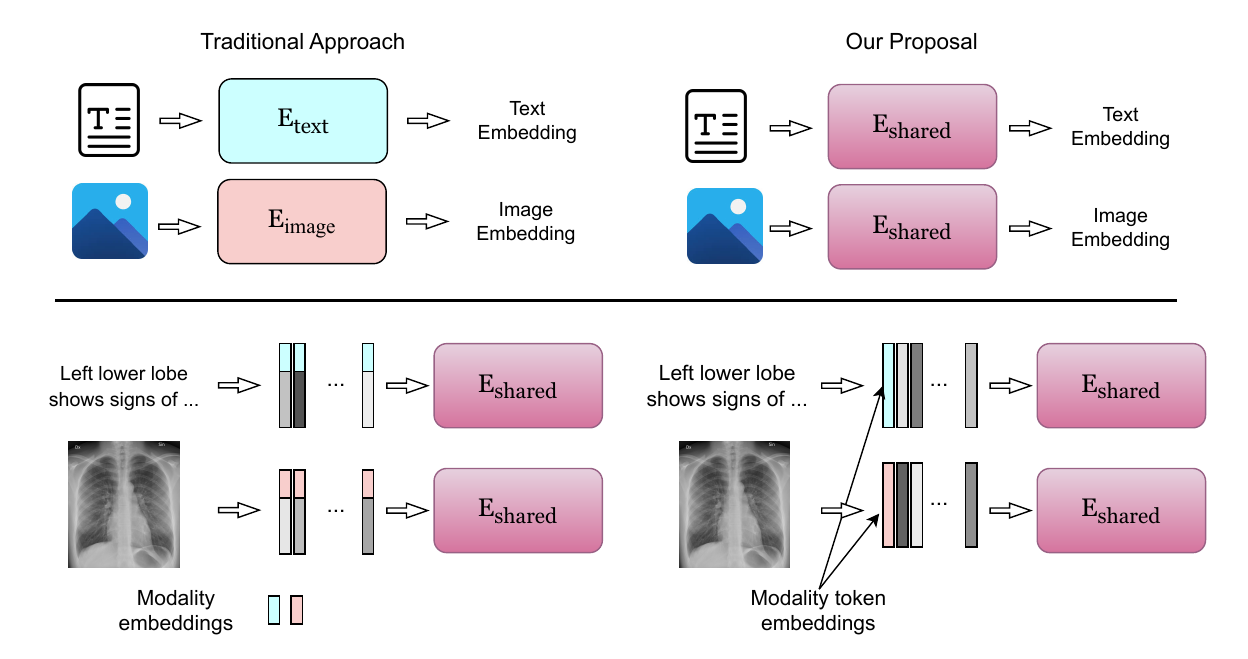}
    \caption{\textbf{Top:} Encoding data with (left) individual encoders vs (right) a shared encoder. \textbf{Bottom:} Adding modality-specific information by (left) appending modality embeddings to feature vectors, vs (right) adding a modality token.}
    \label{fig:main}
\end{figure*}

A fully shared encoder across modalities may struggle to capture modality-specific patterns solely from input sequence embeddings. To improve the model’s ability to learn modality-aware representations, we incorporate explicit modality information into the input. This will help the shared encoder differentiate between modalities while preserving the unified structure. While several strategies may be possible, here we explore two approaches: adding a modality \emph{feature vector} or a modality \emph{token}, explained below.
\begin{itemize}
    \item \emph{Modality Feature Vector.} One approach to incorporate modality information is to append a modality-specific feature vector to all input embedding vectors \cite{lianghigh}. Let $\bv_M$ denote the feature vector for modality $M$. For an input sequence of length $s$ we append $\bv_M$ to all input vectors forming $
    \bh_M^0 = [[\be_M^1; \bv_M], \dots, [\be_M^s; \bv_M]]$.
    \item \emph{Modality Token.} Alternatively, modality information can be represented by a single token attached to the input sequence. Let $\be_M$ denote the embedding vector that corresponds to the modality token for modality M. The input sequence to the encoder will be $\bh_M^0 = [\be_M, \be_M^1, \dots, \be_M^s]$.
\end{itemize}
Figure~\ref{fig:main} (bottom) shows these methods. Adding modality-specific information either in the form of a token or a feature vector ensures that the shared encoder can capture modality-specific nuances while still processing information across all modalities.
      
\subsubsection{Shared Encoder.}
The shared encoder $\shenc$ is a stack of $L$ (transformer) layers that process data from all modalities. The output embedding of the transformer layer can be represented as:
\[
\bh^\ell = \shenc^\ell(\bh^{\ell-1}), \quad \ell \in \{1, \dots, L\},
\]
where $\bh^0 = [\be_M^1,\ldots,\be_M^s]$ represents the input sequence, and $\shenc^\ell$ denotes the $\ell$-th layer of the shared encoder. By sharing the encoder for all modalities the model leverages cross-modal synergies while maintaining computational efficiency. 

\subsection{Layer Sharing and Modality-Specific Layers}
In addition to a fully-shared encoder, we introduce a variant of the model where we allow modality-specific layers in the encoder. While it is possible to add such layers in between the shared layers, for simplicity we only consider modality-specific layers placed either before or after the shared layers:
\begin{equation}
   \mathcal{E} = \psi_M \circ \shenc \circ \phi_M,
   \label{eq:general_form}
\end{equation}
where $\phi_M$ and $\psi_M$ are both modality-specific, $\mathbb{R}^{s \times d} \rightarrow \mathbb{R}^{s \times d}$, sequence-to-sequence encoders.
Early modality-specific layers in particular might be able to capture low-level features specific to the modality.
This hybrid approach balances the need for modality-specific processing and cross-modal alignment, enabling the model to learn both general and specialized features effectively.
We will study the impact of these components in the next section.

\section{Experiments}
\label{sec:experiments}

\subsection{Setup}
To learn representations we employ contrastive learning on image-text pairs as shown in Eq.~\ref{eq:contrastive}. We apply contrastive learning on the representation vector of the [CLS] token for each encoder, which provides a single fixed-length embedding of the entire sequence. We train all encoders on the training split of the PMC-OA dataset \cite{lin2023pmc}, extracted from PubMed Central articles, which contains 1.3M image-text pairs. Retrieval evaluation is performed in-distribution on the test split of PMC-OA, and out-of-distribution on three datasets: MIMIC-CXR \cite{johnson2019mimic} (Radiology), Quilt \cite{ikezogwo2024quilt} (Microscopy), and DeepEyeNet \cite{huang2021deepopht} (Visible Light Photography). We run the experiments using the \mbox{\textbf{mmlearn}} multimodal learning framework \href{https://github.com/VectorInstitute/mmlearn}{(https://github.com/VectorInstitute/mmlearn)}. 

\subsection{Results}
\subsubsection{Comparing Shared vs Modality-Specific Encoders.}
\begin{figure*}[tb]
    \centering
    \includegraphics[width=1.0\linewidth]{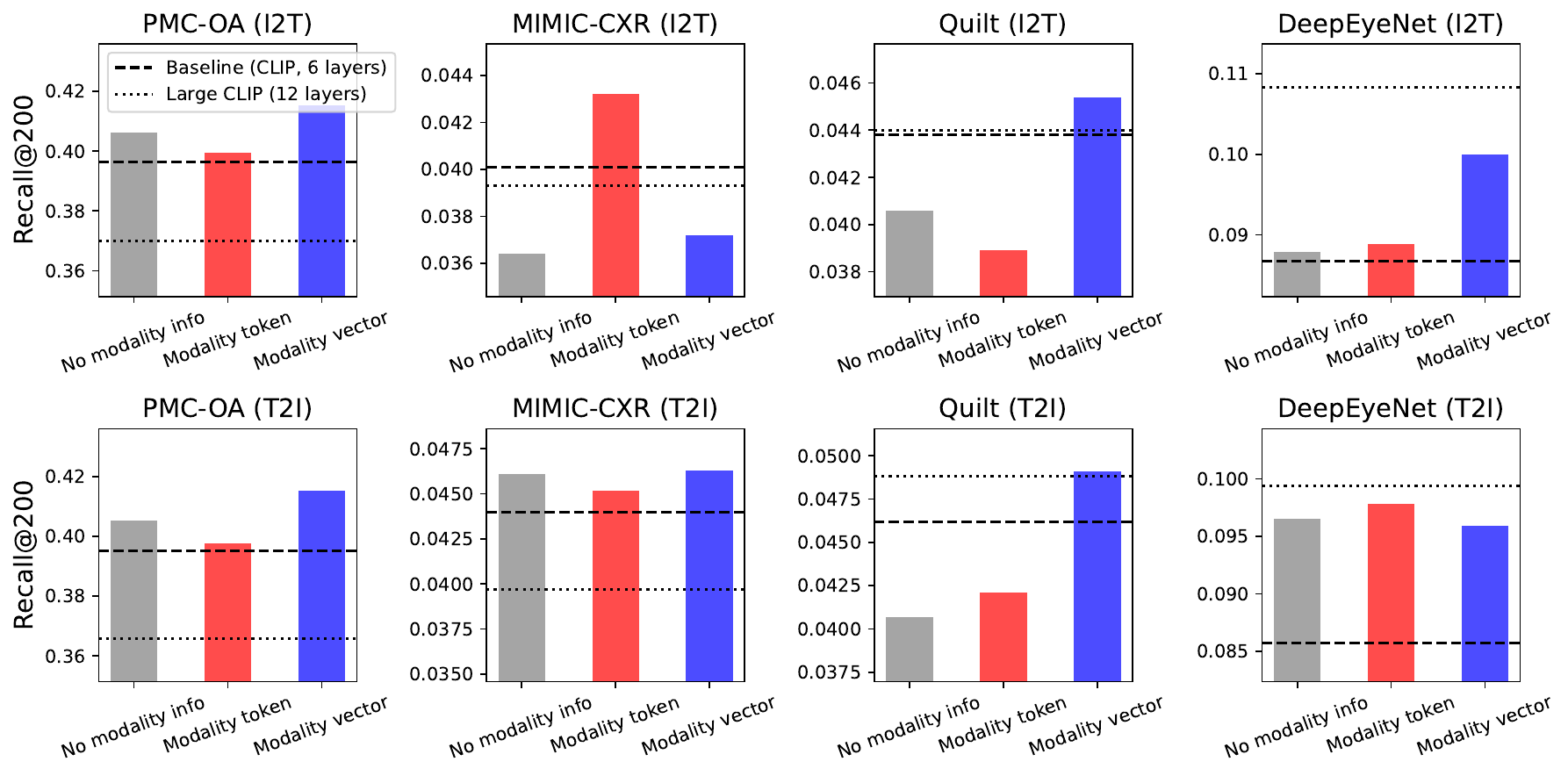}
    \caption{Comparison of methods for incorporating modality information into a shared encoder. The baseline (dark dashed line) uses two disjoint modality-specific encoders (125M parameters total), while the Large CLIP model (dotted line) has 210M parameters. All shared encoders have 125M parameters.}
    \label{fig:main_results}
\end{figure*}

We evaluate shared encoders (Section~\ref{sec:methodology}) against a baseline model with separate image and text encoders (Baseline CLIP). To ensure a fair comparison, the shared encoder matches the total parameters of the modality-specific baseline. The baseline has 6 transformer layers and 125M parameters, so the shared encoder is scaled to 12 layers to match. Additionally, we include a stronger model with modality-specific encoders, each with 12 layers, totaling 210M parameters (Large CLIP).
The shared encoder with \emph{modality vector} allocates 20 of 768 input features to modality-specific information, chosen via grid search over $\{10, 20, 50\}$. Figure~\ref{fig:main_results} presents text-to-image (T2I) and image-to-text (I2T) retrieval results. Among shared encoders, those with modality features outperform alternatives using modality tokens or no modality information in 6 out of 8 tasks, indicating that incorporating modality information improves performance, with modality features being more effective than modality tokens.

Compared to modality-specific models, this model surpasses Baseline CLIP in 7 of 8 tasks and outperforms the Large CLIP model (70\% more parameters) in 5 of 8 tasks, highlighting the benefits of parameter sharing in multimodal representation learning.

\subsubsection{Impact of Parameter Sharing Across Training Set Sizes.}
Parameter sharing across encoders reduces the number of trainable parameters, which could improve generalization particularly in settings with limited training data. We hypothesize that the generalization benefits of parameter sharing are more pronounced in low-data regimes. We test this hypothesis by comparing the performance of shared and modality-specific encoders on PMC-OA across different training set sizes. Table~\ref{tab:train_set_comparison} shows the results. In both Image2Text and Text2Image retrieval tasks, the performance improvement from using a shared encoder over modality-specific encoders is more pronounced with smaller training set sizes.

\begin{table*}[htb]
\centering
\caption{Comparison of modality-specific vs. shared encoders across training set sizes. Both encoders have similar sizes (125M vs. 124M parameters for modality-specific and shared encoder respectively). Numbers indicate Recall@200 for Image2Text and Text2Image retrieval. Relative difference shows the shared encoder's improvement over modality-specific encoders.}
\label{tab:train_set_comparison}
\resizebox{0.8\textwidth}{!}{
\begin{tabular}{@{}l|c|c|c|c|c@{}}
\toprule
\textbf{Task} & \textbf{Model} & \multicolumn{4}{c}{\textbf{Train Set Size (million)}} \\ \cmidrule(lr){3-6}
 & & \textbf{1.74} & \textbf{1.32} & \textbf{0.66} & \textbf{0.33} \\ \midrule
\multirow{3}{*}{Image2Text} & Modality-Specific Encoders & 0.4643 & 0.3965 & 0.1333 & 0.0207  \\
 & Shared Encoder           & 0.4718 & 0.4152 & 0.2413 & 0.0242  \\
 & \textbf{Relative Diff (\%)}       & \textbf{+1.62} & \textbf{+4.72} & \textbf{+81.02} & \textbf{+16.91} \\
 \midrule
\multirow{3}{*}{Text2Image} & Modality-Specific Encoders & 0.4663 & 0.3952 & 0.1220 & 0.0186  \\
                     & Shared Encoder           & 0.4721 & 0.4152 & 0.2369 & 0.0241  \\
                     & \textbf{Relative Diff (\%)}       & \textbf{+1.24} & \textbf{+5.06} & \textbf{+94.18} & \textbf{+29.57} \\ \bottomrule
\end{tabular}
}
\end{table*}

\subsubsection{Integrating Modality-Specific Layers with Shared Encoders.} 
\begin{figure*}[htb]
    \centering
    \includegraphics[width=0.8\linewidth]{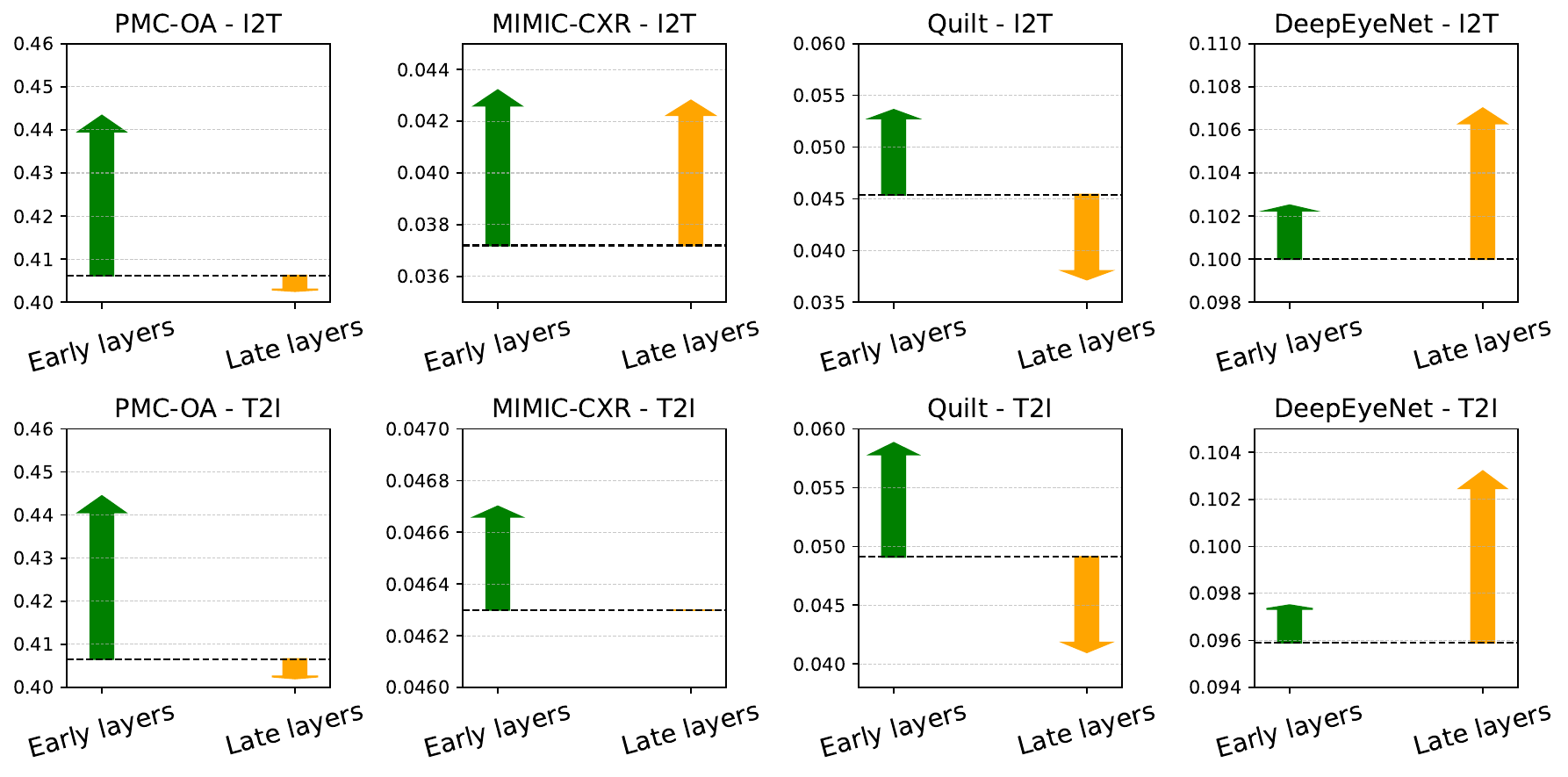}
    \caption{Impact of adding \emph{modality-specific} layers to encoders. The dashed line shows the performance of a purely shared encoder, with green arrows indicating performance gains from early modality layers and orange arrows showing gains (or losses) from late modality layers.}
    \label{fig:mixed_encoder}
\end{figure*}

Next, we examine the impact of combining shared and modality-specific components, as formulated in Eq.~\ref{eq:general_form}. Specifically, we experiment with inserting two modality-specific transformer layers either before or after the shared encoder. As shown in Figure~\ref{fig:mixed_encoder}, incorporating early modality-specific layers consistently improves performance. This suggests that these layers effectively capture low-level, modality-specific features that are crucial for achieving strong performance in each modality. In contrast, adding late modality-specific layers yields mixed results, making it unclear whether their inclusion is beneficial in general.

\FloatBarrier
\section{Conclusion}
We proposed a shared encoder framework for multimodal representation learning in the medical domain, addressing the challenge of limited data. Through extensive experiments we demonstrated significant performance gains with this approach over separate modality-specific encoders, particularly in low-data scenarios. By efficiently leveraging modality information and selective modality-specific layers, our method offers a scalable and effective solution for building multimodal medical models. These results highlight the potential of shared encoder architectures to advance medical AI applications while reducing dependence on large-scale data and proprietary models.
 
\clearpage
\bibliographystyle{splncs04}
\bibliography{ref}

\begin{thebibliography}{10}
\providecommand{\url}[1]{\texttt{#1}}
\providecommand{\urlprefix}{URL }
\providecommand{\doi}[1]{https://doi.org/#1}

\bibitem{baliah2023exploring}
Baliah, S., Maani, F.A., Sanjeev, S., Khan, M.H.: Exploring the transfer learning capabilities of clip in domain generalization for diabetic retinopathy. In: International Workshop on Machine Learning in Medical Imaging. pp. 444--453 (2023)

\bibitem{boecking2022making}
Boecking, B., Usuyama, N., Bannur, S., Castro, D.C., Schwaighofer, A., Hyland, S., Wetscherek, M., Naumann, T., Nori, A., Alvarez-Valle, J., et~al.: Making the most of text semantics to improve biomedical vision--language processing. pp. 1--21. Springer (2022)

\bibitem{chada2023momo}
Chada, R., Zheng, Z., Natarajan, P.: Momo: A shared encoder model for text, image and multi-modal representations. arXiv preprint arXiv:2304.05523  (2023)

\bibitem{eslami2021does}
Eslami, S., de~Melo, G., Meinel, C.: Does clip benefit visual question answering in the medical domain as much as it does in the general domain? arXiv preprint arXiv:2112.13906  (2021)

\bibitem{girdhar2023imagebind}
Girdhar, R., El-Nouby, A., Liu, Z., Singh, M., Alwala, K.V., Joulin, A., Misra, I.: Imagebind: One embedding space to bind them all. In: CVPR. pp. 15180--15190 (2023)

\bibitem{huang2021deepopht}
Huang, J.H., Yang, C.H.H., Liu, F., Tian, M., Liu, Y.C., Wu, T.W., Lin, I.H., Wang, K., Morikawa, H., Chang, H., Tegner, J., Worring, M.: Deepopht: medical report generation for retinal images via deep models and visual explanation. In: WACV. pp. 2442--2452 (2021)

\bibitem{huang2021gloria}
Huang, S.C., Shen, L., Lungren, M.P., Yeung, S.: Gloria: A multimodal global-local representation learning framework for label-efficient medical image recognition. In: CVPR. pp. 3942--3951 (2021)

\bibitem{ikezogwo2024quilt}
Ikezogwo, W., Seyfioglu, S., Ghezloo, F., Geva, D., Sheikh~Mohammed, F., Anand, P.K., Krishna, R., Shapiro, L.: Quilt-1m: One million image-text pairs for histopathology. Advances in Neural Information Processing Systems  \textbf{36} (2024)

\bibitem{ALIGN}
Jia, C., Yang, Y., Xia, Y., Chen, Y.T., Parekh, Z., Pham, H., Le, Q., Sung, Y.H., Li, Z., Duerig, T.: Scaling up visual and vision-language representation learning with noisy text supervision. In: ICML. pp. 4904--4916. PMLR (2021)

\bibitem{jia2021scaling}
Jia, C., Yang, Y., Xia, Y., Chen, Y.T., Parekh, Z., Pham, H., Le, Q., Sung, Y.H., Li, Z., Duerig, T.: Scaling up visual and vision-language representation learning with noisy text supervision. pp. 4904--4916 (2021)

\bibitem{johnson2019mimic}
Johnson, A.E., Pollard, T.J., Berkowitz, S.J., Greenbaum, N.R., Lungren, M.P., Deng, C.y., Mark, R.G., Horng, S.: Mimic-cxr, a de-identified publicly available database of chest radiographs with free-text reports. Scientific Data  \textbf{6}(1), ~317 (2019)

\bibitem{kim2023concept}
Kim, I., Kim, J., Choi, J., Kim, H.J.: Concept bottleneck with visual concept filtering for explainable medical image classification. In: International Conference on Medical Image Computing and Computer-Assisted Intervention. pp. 225--233. Springer (2023)

\bibitem{lei2024vit}
Lei, W., Ge, Y., Yi, K., Zhang, J., Gao, D., Sun, D., Ge, Y., Shan, Y., Shou, M.Z.: Vit-lens: Towards omni-modal representations. In: CVPR. pp. 26647--26657 (2024)

\bibitem{li2021supervision}
Li, Y., Liang, F., Zhao, L., Cui, Y., Ouyang, W., Shao, J., Yu, F., Yan, J.: Supervision exists everywhere: A data efficient contrastive language-image pre-training paradigm. arXiv preprint arXiv:2110.05208  (2021)

\bibitem{lin2023pmc}
Lin, W., Zhao, Z., Zhang, X., Wu, C., Zhang, Y., Wang, Y., Xie, W.: Pmc-clip: Contrastive language-image pre-training using biomedical documents. In: International Conference on Medical Image Computing and Computer-Assisted Intervention. pp. 525--536. Springer (2023)

\bibitem{liu2023imitate}
Liu, C., Cheng, S., Shi, M., Shah, A., Bai, W., Arcucci, R.: Imitate: Clinical prior guided hierarchical vision-language pre-training. arXiv preprint arXiv:2310.07355  (2023)

\bibitem{liu2024etp}
Liu, C., Wan, Z., Cheng, S., Zhang, M., Arcucci, R.: Etp: Learning transferable ecg representations via ecg-text pre-training. In: ICASSP 2024-2024 IEEE International Conference on Acoustics, Speech and Signal Processing (ICASSP). pp. 8230--8234. IEEE (2024)

\bibitem{lu2024visual}
Lu, M.Y., Chen, B., Williamson, D.F., Chen, R.J., Liang, I., Ding, T., Jaume, G., Odintsov, I., Le, L.P., Gerber, G., et~al.: A visual-language foundation model for computational pathology. Nature Medicine  \textbf{30}(3),  863--874 (2024)

\bibitem{lianghigh}
Manna, S., Bhattacharya, S., Pal, U.: Self-supervised visual representation learning for medical image analysis: A comprehensive survey. Transactions on Machine Learning Research  (2024)

\bibitem{muller2022joint}
M{\"u}ller, P., Kaissis, G., Zou, C., Rueckert, D.: Joint learning of localized representations from medical images and reports. In: ECCV. pp. 685--701. Springer (2022)

\bibitem{clip}
Radford, A., Kim, J.W., Hallacy, C., Ramesh, A., Goh, G., Agarwal, S., Sastry, G., Askell, A., Mishkin, P., Clark, J., et~al.: Learning transferable visual models from natural language supervision. In: ICML. pp. 8748--8763 (2021)

\bibitem{radford2021learning}
Radford, A., Kim, J.W., Hallacy, C., Ramesh, A., Goh, G., Agarwal, S., Sastry, G., Askell, A., Mishkin, P., Clark, J., et~al.: Learning transferable visual models from natural language supervision. In: ICML. pp. 8748--8763 (2021)

\bibitem{reedgeneralist}
Reed, S., Zolna, K., Parisotto, E., Colmenarejo, S.G., Novikov, A., Barth-maron, G., Gim{\'e}nez, M., Sulsky, Y., Kay, J., Springenberg, J.T., et~al.: A generalist agent. Transactions on Machine Learning Research

\bibitem{seibold2022breaking}
Seibold, C., Rei{\ss}, S., Sarfraz, M.S., Stiefelhagen, R., Kleesiek, J.: Breaking with fixed set pathology recognition through report-guided contrastive training. In: International Conference on Medical Image Computing and Computer-Assisted Intervention. pp. 690--700. Springer (2022)

\bibitem{singh2022flava}
Singh, A., Hu, R., Goswami, V., Couairon, G., Galuba, W., Rohrbach, M., Kiela, D.: Flava: A foundational language and vision alignment model. In: CVPR. pp. 15638--15650 (2022)

\bibitem{tiu2022expert}
Tiu, E., Talius, E., Patel, P., Langlotz, C.P., Ng, A.Y., Rajpurkar, P.: Expert-level detection of pathologies from unannotated chest x-ray images via self-supervised learning. Nature Biomedical Engineering  \textbf{6}(12),  1399--1406 (2022)

\bibitem{wang2305one}
Wang, P., Wang, S., Lin, J., Bai, S., Zhou, X., Zhou, J., Wang, X., Zhou, C.: One-peace: Exploring one general representation model toward unlimited modalities. arxiv 2023. arXiv preprint arXiv:2305.11172

\bibitem{zhai2022lit}
Zhai, X., Wang, X., Mustafa, B., Steiner, A., Keysers, D., Kolesnikov, A., Beyer, L.: Lit: Zero-shot transfer with locked-image text tuning. In: CVPR. pp. 18123--18133 (2022)

\bibitem{zhang2023biomedclip}
Zhang, S., Xu, Y., Usuyama, N., Xu, H., Bagga, J., Tinn, R., Preston, S., Rao, R., Wei, M., Valluri, N., et~al.: Biomedclip: a multimodal biomedical foundation model pretrained from fifteen million scientific image-text pairs. arXiv preprint arXiv:2303.00915  (2023)

\bibitem{zhang2023text}
Zhang, Y., Gao, J., Zhou, M., Wang, X., Qiao, Y., Zhang, S., Wang, D.: Text-guided foundation model adaptation for pathological image classification. In: International Conference on Medical Image Computing and Computer-Assisted Intervention. pp. 272--282. Springer (2023)

\end{thebibliography}

\clearpage

\renewcommand{\thesection}{S\arabic{section}}
\renewcommand{\thefigure}{S\arabic{figure}}
\renewcommand{\thetable}{S\arabic{table}}

\setcounter{table}{0}
\setcounter{figure}{0}
\setcounter{section}{0}
\setcounter{page}{1}

% end of commenting out the appendix

\end{document}